\PassOptionsToPackage{table}{xcolor}
\documentclass[sigconf]{acmart}
\usepackage{colortbl}   
\AtBeginDocument{%
  }
    
\usepackage{multirow}
\usepackage{float}
\usepackage[most]{tcolorbox}
\usepackage{listings}
\usepackage{hyperref}
\usepackage{subcaption}
\usepackage{amsfonts}
\usepackage{marvosym}
\newcommand{\corrmark}{\texorpdfstring{\textsuperscript{\Letter}}{}}

\copyrightyear{2025}
\acmYear{2025}
\setcopyright{cc}
\setcctype{by-nc-sa}
\acmConference[MM '25]{Proceedings of the 33rd ACM International Conference on Multimedia}{October 27--31, 2025}{Dublin, Ireland}
\acmBooktitle{Proceedings of the 33rd ACM International Conference on Multimedia (MM '25), October 27--31, 2025, Dublin,
Ireland}\acmDOI{10.1145/3746027.3758295}
\acmISBN{979-8-4007-2035-2/2025/10}
\thanks{This work is completed during Jieyu Li's internship at A*STAR}
\thanks{\corrmark Corresponding author: Joey Tianyi Zhou.}

\begin{document}

\title{
\raisebox{-0.25em}{\includegraphics[height=1.4em]{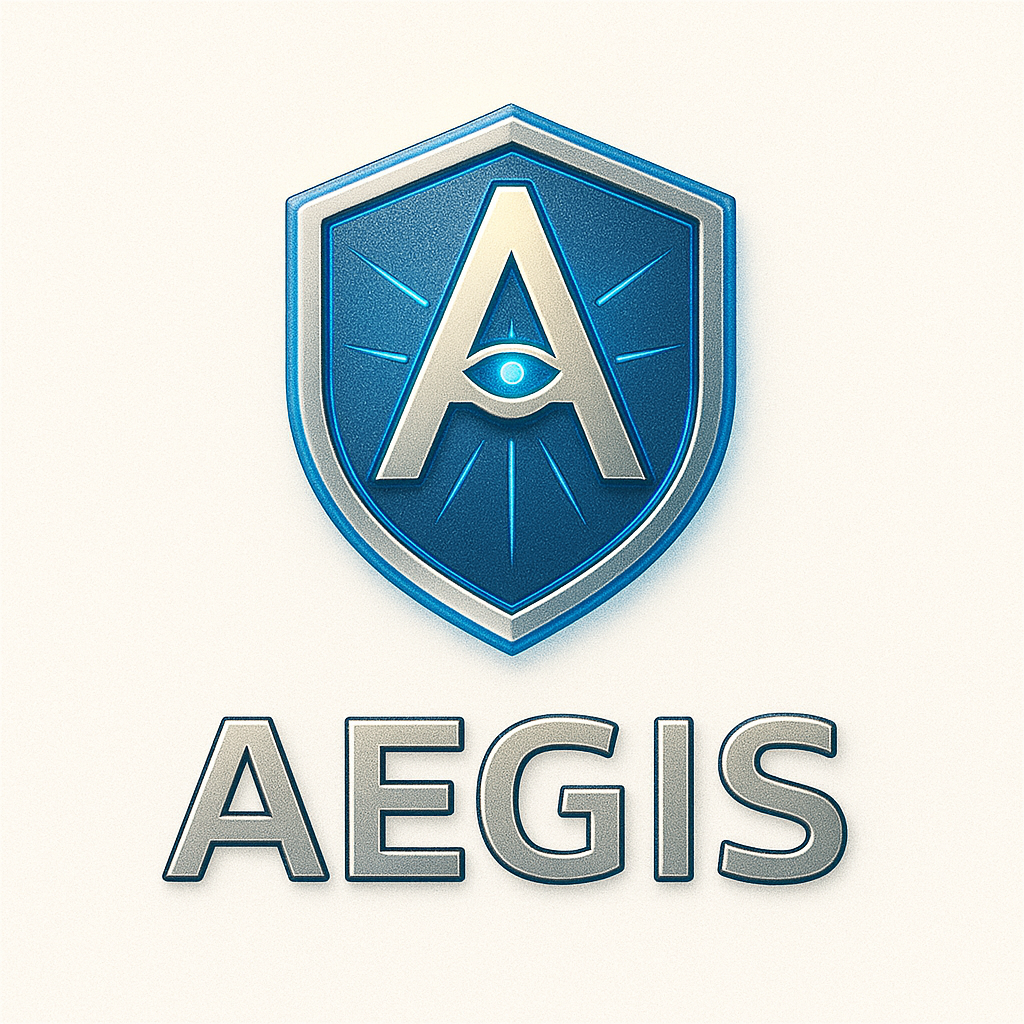}}
{AEGIS: Authenticity Evaluation Benchmark for AI-Generated Video Sequences}
}
\author{Jieyu Li}
\affiliation{%
  \institution{\small National University of Singapore}
  \country{Singapore}
}
\email{jieyuli@u.nus.edu}

\author{Xin Zhang}
\affiliation{%
  \department{\small Centre for Frontier AI Research, \\Institute of High Performance Computing}
  \institution{\small Agency for Science, Technology and Research}
  \country{Singapore}
}
\email{zhangx7@cfar.a-star.edu.sg}

\author{Joey Tianyi Zhou\corrmark}
\affiliation{%
  \department{\small Centre for Frontier AI Research, \\Institute of High Performance Computing}
  \institution{\small Agency for Science, Technology and Research}
  \country{Singapore}
}
\email{joey_zhou@cfar.a-star.edu.sg}

\renewcommand{\shortauthors}{Jieyu Li, Xin Zhang, and Joey Tianyi Zhou}

\begin{abstract}
Recent advances in AI-generated content have fueled the rise of highly realistic synthetic videos, posing severe risks to societal trust and digital integrity.
Existing benchmarks for video authenticity detection typically suffer from limited realism, insufficient scale, and inadequate complexity, failing to effectively evaluate modern vision-language models against sophisticated forgeries. 
To address this critical gap, we introduce \textbf{AEGIS}, a novel large-scale benchmark explicitly targeting the detection of \textit{\textbf{hyper-realistic}} and \textbf{\textit{semantically nuanced}} AI-generated videos. 
AEGIS comprises over 10,000 rigorously curated real and synthetic videos generated by diverse, state-of-the-art generative models, including Stable Video Diffusion, CogVideoX-5B, KLing, and Sora, encompassing open-source and proprietary architectures. 
In particular, AEGIS features specially constructed challenging subsets enhanced with GPT-4o-refined prompts, creating unprecedentedly realistic scenarios for rigorous robustness evaluation. \textcolor{black}{Furthermore, we provide multimodal annotations spanning \textbf{\textit{Semantic-Authenticity Descriptions}}, \textbf{\textit{Motion Features}}, and \textbf{\textit{Low-level Visual Features}}, facilitating authenticity detection and supporting downstream tasks such as multimodal fusion and forgery localization.}
Extensive experiments using advanced vision-language models demonstrate limited detection capabilities on the most challenging subsets of AEGIS, highlighting the dataset’s unique complexity and realism beyond the current generalization capabilities of existing models.
In essence, AEGIS establishes an indispensable evaluation benchmark, fundamentally advancing research toward developing \textit{\textbf{genuinely robust}}, \textit{\textbf{reliable}}, and \textit{\textbf{broadly generalizable}} video authenticity detection methodologies capable of addressing real-world forgery threats.
Our dataset is avaliable on \href{huggingface}{https://huggingface.co/datasets/Clarifiedfish/AEGIS}.
\end{abstract}


\begin{teaserfigure}
  \vspace{-0.5em}
  \includegraphics[width=\textwidth]{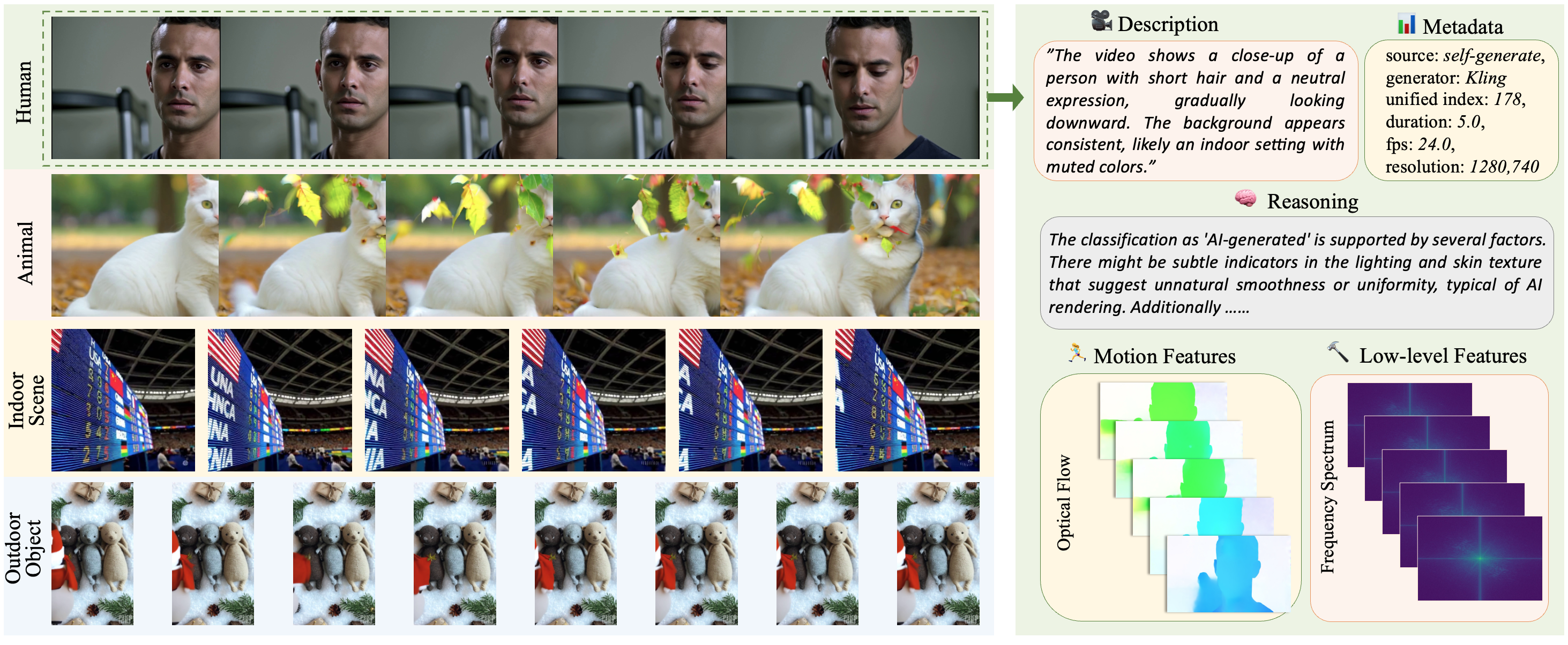}
  \vspace{-2.5em}
  \caption{Overview of the AEGIS dataset. AEGIS comprises a diverse collection of synthetic videos covering a broad range of realistic scenarios, including detailed human facial expressions, natural outdoor animal behaviors, indoor public environments, and intricate static object arrangements. 
  Each video is accompanied by rich multimodal annotations, including \textbf{\textit{Semantic-Authenticity Descriptions}} (semantic summaries, generation metadata, and reasoning explanations), \textbf{\textit{Motion Features}}, and \textbf{\textit{Low-level Visual Features}}. These annotations enable robust and explainable analysis, supporting research in video authenticity detection as well as a variety of downstream multimodal understanding tasks.
}
  \label{fig:features}
\end{teaserfigure}
\received{30 May 2025}
\received[accepted]{1 Aug 2025}

\maketitle
\section{Introduction}
Recent advances in AI-generated content (AIGC) technologies have substantially simplified the creation of highly realistic video content ~\cite{liu2024survey_arxiv, lin2024detecting_arxiv, liu2024sora_arxiv2024}. 
These sophisticated generative models have significantly reduced production costs and fostered novel applications in various fields, such as education and entertainment \cite{zhang2025generative_arxiv2025}. 
However, this rapid proliferation also poses substantial social risks ~\cite{hwang2021effects_cyberpsychology2021, jin2025assessing_nms2025}. 
Compared to synthetic images, the higher perceptual realism and temporal consistency inherent in generated videos exacerbate their potential for misinformation dissemination~\cite{vaccari2020deepfakes_sms2020}, erosion of public trust, and threats to information security across social and professional platforms. 
Consequently, there is an urgent need for robust and reliable detection methods capable of effectively distinguishing synthetic from authentic videos.

However, the advancement of robust video forgery detection methods critically depends on suitable benchmarks. 
Although several AIGC datasets have recently been proposed for forgery detection, they primarily target static images~\cite{wang2023aigciqa2023, li2023agiqa_tcvst2023, li2024aigiqa_cvpr2024}, and thus inherently fail to capture video-specific challenges such as temporal coherence, realistic motion dynamics, and semantic consistency across frames.
Recent video benchmarks, such as VBench~\cite{huang2024vbench_cvpr2024}, EvalCrafter~\cite{liu2024evalcrafter_cvpr2024}, and AIGCBench~\cite{fan2023aigcbench_arxiv2023}, primarily focus on evaluating generation quality or perceptual fidelity rather than explicitly targeting authenticity detection tasks.
Additionally, dedicated video forgery datasets like GenVidBench~\cite{ni2025genvidbench_arxiv2025} and DeMamba~\cite{chen2024demamba_arxiv2024} also possess certain constraints, including simple animation-oriented content, narrow generative diversity, and insufficient emphasis on realism and detection complexity, thereby restricting their capability to comprehensively assess advanced detection algorithms.

To effectively overcome these critical limitations, we propose \textbf{AEGIS}, \textbf{A}uthenticity \textbf{E}valuation Benchmark for AI-\textbf{G}enerated v\textbf{I}deo \textbf{S}equences, a novel video authenticity benchmark meticulously designed to challenge and advance current detection capabilities against highly deceptive AI-generated content. 
AEGIS sets itself apart by exclusively assembling {5,199} synthetic videos derived from seven cutting-edge generative techniques, including prominent open-source methods such as \textcolor{black}{Stable Video Diffusion~\cite{blattmann2023stable_arxiv2023}, CogVideoX-5B~\cite{yang2024cogvideox_arxiv2024}, and I2VGen-XL~\cite{zhang2023i2vgenxl_arxiv2023}, as well as proprietary commercial systems represented by KLing~\cite{kling2024_website}, Sora~\cite{sora2024_website}, and Pika~\cite{pika2024_website}.}
The distinctive integration of these diverse generative techniques ensures unmatched realism, complexity, and representation of current generation paradigms, positioning AEGIS as an indispensable resource for rigorously evaluating and substantially improving model robustness in the face of emerging, highly realistic forgery threats.

To rigorously benchmark and substantially advance video forgery detection, AEGIS introduces several critical innovations that clearly distinguish it from existing benchmarks.
First, it includes carefully crafted challenging subsets refined via GPT-4o~\cite{openai2024gpt4o}-generated prompts, explicitly designed to intensively evaluate model robustness against highly sophisticated and semantically nuanced forgeries.
These synthetic videos are complemented by systematically curated authentic videos characterized by significant visual complexity and diversity, creating realistic evaluation conditions.
Second, AEGIS provides extensive multimodal annotations, including optical flow, frequency-domain analysis, and rich semantic descriptions, to facilitate rigorous evaluation and support diverse downstream forensic tasks.
Crucially, our extensive experiments using SOTA vision-language models, such as Qwen-VL~\cite{bai2025qwen2.5-vl_arxiv2025} and Video-LLaVA ~\cite{lin2023videollava_arxiv2023}, across zero-shot, prompt-guided, and fine-tuned scenarios, reveal notable performance gaps, especially within challenging subsets.
These findings underscore both the substantial difficulty presented by AEGIS and its effectiveness to expose critical generalization limitations inherent in current detection methodologies.
Consequently, AEGIS emerges as an indispensable benchmark, uniquely positioned to drive forward the development of more robust and widely generalizable video authenticity detection models.

The key contributions of this work include:
\begin{itemize}
\item We propose \textbf{AEGIS}, a novel large-scale benchmark for video authenticity detection, comprising 5,199 synthetic videos generated by six diverse SOTA techniques, covering both open-source and proprietary models. AEGIS significantly improves upon existing benchmarks in diversity, realism, and semantic complexity.
\item We design challenging evaluation subsets using GPT-4o-refined prompts to simulate highly realistic, semantically nuanced scenarios. The resulting \textit{Hard Test Sets} effectively expose generalization gaps in current detection models.
\item We curate authentic videos with rich multimodal annotations, including semantic descriptions, optical flow, frequency-domain features, and temporal coherence metrics. Experiments with advanced vision-language models reveal clear performance limitations, underscoring AEGIS’s value for robust and generalizable forgery detection.
\end{itemize}

\begin{table*}[t!]
\footnotesize\centering
\caption{Overview of the Collected Dataset}
\vspace{-1.3em}
\label{tab:dataset-overview}
\begin{tabular}{ccccccccc}
  \hline
  \rowcolor[HTML]{EFEFEF} 
  {\color[HTML]{333333} \textbf{Split}}   & \multicolumn{1}{c}{{\color[HTML]{333333} \textbf{\begin{tabular}[c]{@{}c@{}}Total  number\end{tabular}}}} & \multicolumn{1}{c}{{\color[HTML]{333333} \textbf{Category}}} & \multicolumn{1}{c}{{\color[HTML]{333333} \textbf{Number}}} & \multicolumn{1}{c}{{\color[HTML]{333333} \textbf{Source}}} & \multicolumn{1}{c}{{\color[HTML]{333333} \textbf{Duration (s)}}} & \multicolumn{1}{c}{{\color[HTML]{333333} \textbf{Number}}} & {\color[HTML]{333333} \textbf{Frame rate (fps)}} & \multicolumn{1}{c}{{\color[HTML]{333333} \textbf{Resolution}}} \\ \hline
& &&& TIP-I2V (Pika)&& 756& 24 &  \\
& &&& TIP-I2V (CogVideoX-5B) && 886& 8  &  \\
& &&& TIP-I2V (Stable Video Diffusion) && 621& 7  &  \\
& & \multirow{-4}{*}{Synthetic}& \multirow{-4}{*}{3161} & TIP-I2V (I2VGen-XL)  & \multirow{-4}{*}{3-7} & 898& 7  &  \\ \cline{3-8}
& &&& Vript (YouTube)  && 4060      &    &  \\
  \multirow{-6}{*}{Traing Set}& \multirow{-6}{*}{7304}& \multirow{-2}{*}{Real}  & \multirow{-2}{*}{4143} & Vript (TikTok) & \multirow{-2}{*}{5-10}& 283& \multirow{-2}{*}{24-30} &  \\ \cline{1-8}
      && && TIP-I2V (Pika)& & 455    & 24          &  \\
      && && TIP-I2V (CogVideoX-5B) & & 455    & 8&  \\
      && && TIP-I2V (Stable Video Diffusion) & & 455    & 7&  \\
      && \multirow{-4}{*}{Synthetic} & \multirow{-4}{*}{1820}         & TIP-I2V (I2VGen-XL)      & \multirow{-4}{*}{3-7}  & 455    & 7&  \\ \cline{3-8}
      && && Vript (YouTube)  & & 455    & &  \\
  \multirow{-6}{*}{Validation Set}        & \multirow{-6}{*}{2730}& \multirow{-2}{*}{Real}   & \multirow{-2}{*}{910}          & Vript (TikTok) & \multirow{-2}{*}{5-10} & 455    & \multirow{-2}{*}{24-30} & \multirow{-12}{*}{380p-1080p}  \\ \hline
&& && Sora    & & 107    & 30          &  \\
&& \multirow{-2}{*}{Synthetic} & \multirow{-2}{*}{218}          & KLing   & \multirow{-2}{*}{5s}     & 111    & 24          &  \\ \cline{3-8}
&& && DVF& & 109    & &  \\ 
  \multirow{-4}{*}{Hard Test Set}       & \multirow{-4}{*}{436} & \multirow{-2}{*}{Real}   & \multirow{-2}{*}{218}          & self-collected (YouTube)& \multirow{-2}{*}{2.4-10}& 109    & \multirow{-2}{*}{24-30} & \multirow{-4}{*}{270p-4K} \\ \hline
  \end{tabular}
\vspace{-1em}
\end{table*}
\section{Related Work}
\subsection{Image-level AIGC Benchmark}
Synthetic image generation has significantly advanced due to Generative Adversarial Networks (GANs)~\cite{goodfellow2014generative_nips2014}, diffusion models~\cite{croitoru2023diffusion_tpami2023}, and flow-matching techniques~\cite{klein2023equivariant_nips2023,gat2024discrete_nips2024}. 
Early models such as StyleGAN~\cite{karras2020analyzing_cvpr2020} notably enhanced facial realism, while recent diffusion-based and transformer-based approaches, including Stable Diffusion and DALLE-2, significantly expanded general-purpose image synthesis~\cite{zhou2023shifted_cvpr2023,li2024stylegan_cvpr2024}.
Correspondingly, several benchmarks, including AIGCIQA2023~\cite{wang2023aigciqa2023}, AGIQA-20K~\cite{li2024aigiqa_cvpr2024}, PKU-AIGIQA-4K~\cite{yuan2024pku_arxiv2024}, and FragFake~\cite{sun2025fragfakedatasetfinegraineddetection_arxiv2025}, emerged to rigorously evaluate image generation quality, focusing primarily on perceptual fidelity, semantic alignment, and fine-grained detection tasks. 
Despite these advancements, these image-level datasets inherently lack consideration of video-specific challenges such as temporal coherence and realistic motion patterns. 
Our proposed AEGIS explicitly addresses these critical video-centric issues by integrating temporal and multimodal analysis, significantly extending beyond static imagery benchmarks.

\subsection{Video-level AIGC Benchmarks}
Recent video-level AI-generated content (AIGC) benchmarks such as VBench~\cite{huang2024vbench_cvpr2024}, T2VSafetyBench~\cite{miao2024t2vsafetybench_nips2024}, EvalCrafter~\cite{liu2024evalcrafter_cvpr2024}, and VIDEOPHY~\cite{bansal2024videophy_arxiv2024}, primarily focus on assessing video generation quality rather than explicitly addressing authenticity detection tasks.
Meanwhile, existing datasets specifically targeting video forgery detection, such as DF40~\cite{yan2024df40_arxiv2024}, Deepfake-Eval-2024~\cite{chandra2025deepfake_arxiv2025}, and ExDDV~\cite{hondru2025exddv_arxiv2025}, predominantly emphasize facial manipulation and in-the-wild deepfakes, limiting their broader generalization. 
Recent large-scale benchmarks like GenVidBench~\cite{ni2025genvidbench_arxiv2025} and DeMamba~\cite{chen2024demamba_arxiv2024}, while including diverse generative sources, often incorporate less challenging animation-style videos or emphasize dataset scale over detection complexity.
In contrast, our proposed AEGIS benchmark explicitly prioritizes video authenticity detection by focusing solely on hyper-realistic, semantically complex AI-generated videos, deliberately excluding easily identifiable animation-oriented content. 
Unlike prior datasets, AEGIS provides detailed multimodal annotations and includes specially constructed challenging subsets enhanced by GPT-4o-refined prompts, explicitly designed to rigorously evaluate the robustness and generalization capabilities of detection methods.
\section{Dataset Construction}
\begin{figure*}
    \centering
    \includegraphics[width=\linewidth]{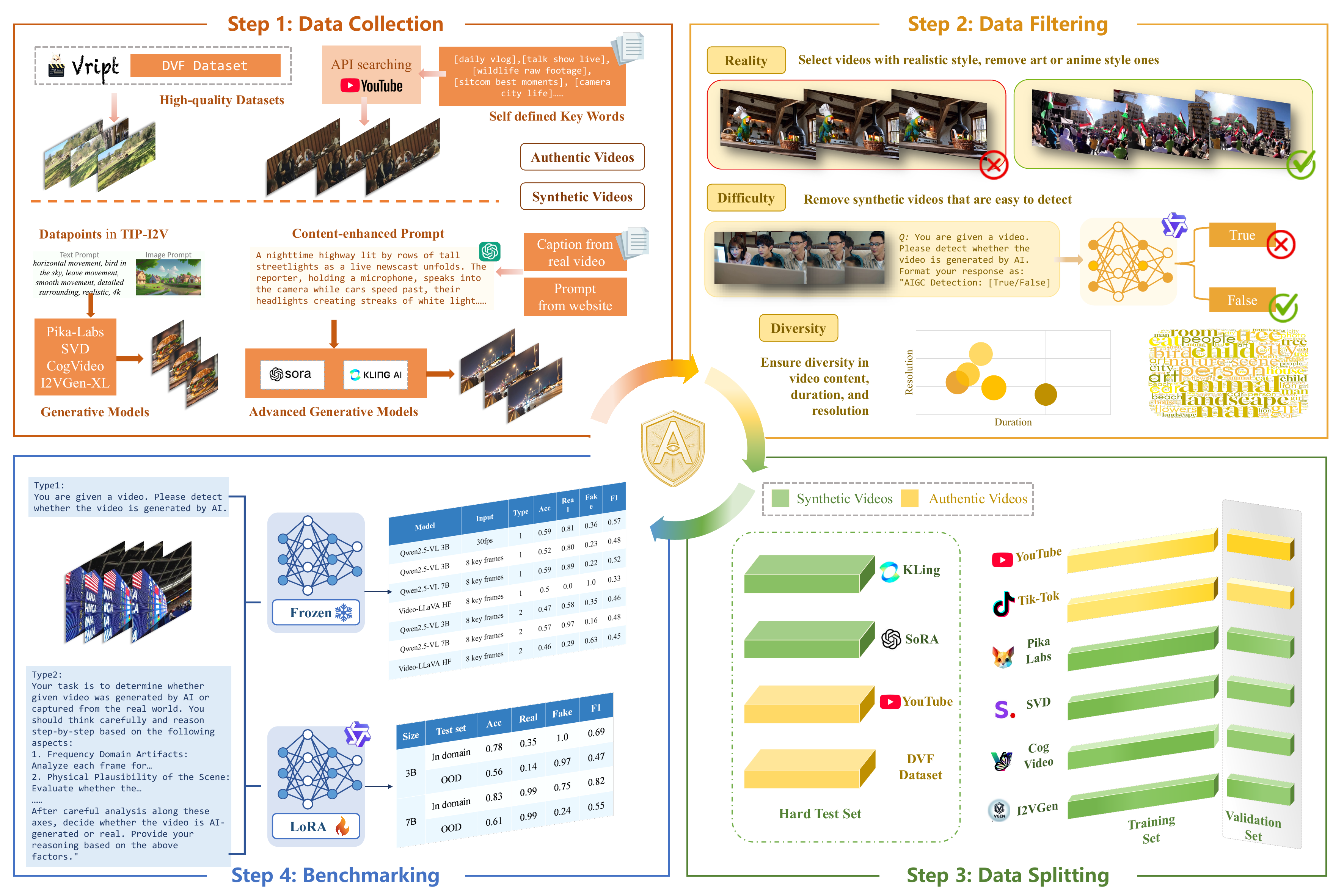}
    \caption{The AEGIS Dataset Construction Pipeline. Step 1: Data Collection – Collecting real and synthetic videos from diverse sources.
Step 2: Data Filtering – Applying three key principles: \textbf{reality} (removing non-photorealistic content), \textbf{difficulty} (discarding easily detectable fakes), and \textbf{diversity} (ensuring variation in content, resolution, and duration).
Step 3: Data Splitting – Creating balanced training, validation, and hard test sets.
Step 4: Benchmarking – Evaluating vision-language models under different settings including zero-shot inference, structured reasoning prompt, and low-rank adaptation (LoRA) fine-tuning.\protect\footnotemark}
    \label{fig:construction pipeline}
\end{figure*}

To advance research in video authenticity detection, we construct AEGIS, a comprehensive dataset consisting of both AI-generated and real-world videos.
This section details our systematic construction pipeline, as illustrated in \autoref{fig:construction pipeline}, which comprises three main stages: data collection, data filtering, and data splitting.
An overview of the final structure of the AEGIS dataset is provided in \autoref{tab:dataset-overview}.
\footnotetext{Parts of the video frame examples in \autoref{fig:features} and \autoref{fig:construction pipeline} are sourced from \cite{wang2024tip_arxiv2024} and \cite{yang2024vript_nips2024}.}
\subsection{Data Collection}
\subsubsection{Real Video Collection}
To ensure high realism, diversity, and visual complexity, we collect authentic videos from three sources:
(1) \textbf{Vript Dataset}~\cite{yang2024vript_nips2024}: We utilize approximately 12,000 annotated videos sourced from YouTube (horizontal, long-form) and TikTok (vertical, short-form). 
This cross-platform selection captures inherent content and stylistic biases associated with different formats, 
thus preserving the diversity characteristic of real-world videos.
(2) \textbf{DVF Dataset}~\cite{song2024onlearning_arxiv2025}: Specifically selected for its high visual complexity and realism, DVF consists of diverse human-recorded video clips that capture subtle details and closely mimic the complexities of genuine human-generated content.
(3) \textbf{Supplemental YouTube Collection}: To further augment the realism and practical applicability of our dataset, we independently collected minimally edited videos from YouTube, including raw street interviews, wildlife documentaries, and authentic vlogs. 
We systematically applied 30 predefined search queries designed to maximize diversity and ensure authenticity. 
Each collected video was rigorously verified through manual inspection and standardized by trimming clips to durations ranging from 2.4 to 10 seconds, removing audio tracks, and maintaining a resolution diversity spanning from 360p to 4K. 
This careful standardization process significantly improves the dataset's realism, representativeness, and complexity, effectively facilitating robust model evaluation.

\subsubsection{Synthetic Video Collection}
Our synthetic subset rigorously integrates publicly available advanced datasets and independently generated synthetic content, employing SOTA generative models to ensure both diversity and realistic complexity.  
(1) \textbf{TIP-I2V Dataset}~\cite{wang2024tip_arxiv2024} \footnote{We used the official curated subset of the original TIP-I2V dataset, available at \url{https://huggingface.co/datasets/WenhaoWang/TIP-I2V/tree/main/subset_videos_tar}}: This dataset provides 500,000 synthetic video clips, generated from 100,000 prompts using five SOTA video generation models: Stable Video Diffusion~\cite{blattmann2023stable_arxiv2023}, CogVideoX-5B~\cite{yang2024cogvideox_arxiv2024}, I2VGen-XL~\cite{zhang2023i2vgenxl_arxiv2023}, Open-Sora~\cite{zheng2024opensora_arxiv2024}, and Pika~\cite{pika2024_website}. 
(2) \textbf{Proprietary Model Generation via KLing and Sora}: To further evaluate and enhance model robustness under challenging scenarios, we independently generated synthetic videos utilizing proprietary, SOTA generative models, namely KLing~\cite{kling2024_website} and Sora~\cite{sora2024_website}. 
High-quality textual prompts were meticulously sourced from the extensive HD-VG-130M dataset~\cite{wang2023videofactory_arxiv2023},
and carefully crafted with reference to exemplary demonstrations provided by the official KLing ~\cite{kling2024_website} and Sora ~\cite{sora2024_website} showcases. 
Each prompt underwent an additional refinement process leveraging GPT-4o ~\cite{openai2024gpt4o}, ensuring enhanced detail, semantic precision, and realism. 
We systematically generated a balanced collection of 218 synthetic videos, carefully controlling for diversity in visual content, resolutions ranging from 360p to 1080p, and durations varying from 5 to 10 seconds.

\subsection{Data Filtering}
To support robust training and evaluation of authenticity detection models, we propose a unified filtering framework grounded in three principles, \textbf{\textit{Reality}}, \textbf{\textit{Difficulty}}, and \textbf{\textit{Diversity}}, applied to both real and synthetic subsets. The framework ensures that real-world samples exhibit high-fidelity, unaltered content, while synthetic samples are photorealistic yet non-trivial to detect. It further promotes content diversity critical for generalization across visually complex scenarios.

(1) \textbf{\textit{Reality}.} We ensured that all included videos—whether real or synthetic—exhibit photorealistic styles by removing art-style, low-quality, or heavily edited content.
For the real videos, since datasets like Vript~\cite{yang2024vript_nips2024} and the Supplemental YouTube Collection contain web-sourced material, many clips were either AI-generated or excessively edited. To address this, we employed Qwen2.5-VL~\cite{bai2025qwen2.5-vl_arxiv2025} to classify approximately 9,000 clips into camera-shot (authentic), heavily edited, and AI-generated categories, discarding around 4,000 non-authentic samples. For the DVF dataset, we performed additional manual reviews to ensure visual quality and authenticity.
For the synthetic videos from the TIP-I2V dataset~\cite{wang2024tip_arxiv2024}, we first selected high-quality, photorealistic prompts from the 100,000 officially provided. Prompts with abstract, implausible, or stylistically inconsistent content were excluded. We then verified that the described actions were visually feasible and semantically coherent, filtering out physically unrealistic or static scenarios. After this two-stage filtering process, 17,000 prompts were retained. For each prompt, we randomly selected one generated video from among five aforementioned candidate models. Due to the relatively low visual quality of Open-Sora’s outputs~\cite{zheng2024opensora_arxiv2024}, its generated videos were further removed from our subset.
In addition, we applied manual filtering to ensure the visual realism of videos generated by KLing~\cite{kling2024_website} and Sora~\cite{sora2024_website}.

(2) \textbf{\textit{Difficulty}.} After the Reality filtering, the remaining synthetic videos from the TIP-I2V dataset underwent an additional round of screening using zero-shot classification with the Qwen2.5-VL model~\cite{bai2025qwen2.5-vl_arxiv2025}. This step was designed to further eliminate samples that exhibit obvious signs of synthetic generation. Specifically, we used Qwen2.5-VL to classify each video as either “AI-generated” (True) or “not AI-generated” (False). Videos that were confidently predicted as “AI-generated” were discarded. In contrast, videos labeled as “not AI-generated” with lower model confidence were retained. This procedure resulted in a curated subset of approximately 5,000 synthetic clips, which exhibit greater visual realism and are less likely to be trivially detected as generated. This additional refinement step ensures that the synthetic data used in our evaluation is both challenging and of high perceptual quality.

(3) \textbf{\textit{Diversity}.} We ensured diversity in both real and synthetic videos across scene types, subjects, and visual conditions.
Real clips span indoor and outdoor environments, human, animal, and object subjects, as well as urban and rural scenes, with durations ranging from 2.4 to 10 seconds and resolutions from 360p to 4K.
Synthetic videos were generated from diverse prompts using four SOTA generators, introducing variations in motion, actors, backgrounds, and lighting conditions. Scene tag distributions (visualized via a word cloud) and resolution-duration scatter plots confirm broad coverage across both semantic and visual dimensions.

Overall, our comprehensive filtering pipeline ensures that the resulting dataset is both high-quality and representative, enabling robust and realistic evaluation of video authenticity detection methods. After applying all filtering procedures, the finalized AEGIS dataset comprises approximately 5,199 synthetic and 5,271 authentic videos, systematically curated to support reliable benchmarking across diverse and challenging scenarios.

\subsection{Data Splitting}
\label{sec:data_split}
To effectively benchmark models under realistic deployment scenarios and rigorously evaluate their generalization capabilities, we systematically divided the filtered AEGIS dataset into three subsets: {Training Set}, {Validation Set}, and {Hard Test Set}. 
The Training and Validation Sets primarily include filtered authentic videos from Vript dataset~\cite{yang2024vript_nips2024} and high-quality synthetic videos from TIP-I2V dataset ~\cite{wang2024tip_arxiv2024}.
The Training Set facilitates the learning of discriminative features that distinguish authentic from synthetic videos, while the Validation Set supports hyperparameter tuning and preliminary model evaluation.
The Hard Test Set specifically evaluates model robustness and generalization under more challenging conditions. It comprises diverse authentic videos from DVF dataset ~\cite{song2024onlearning_arxiv2025} and Supplemental YouTube Collection, and advanced synthetic videos generated by proprietary models KLing~\cite{kling2024_website} and Sora ~\cite{sora2024_website}. Selected for complexity and subtlety, these samples provide a critical benchmark for assessing models' capabilities in realistic scenarios involving sophisticated forgeries and nuanced visual details.

\subsection{Multimodal Annotations}
Effectively distinguishing AI-generated videos from authentic ones requires capturing and representing complementary visual cues across multiple dimensions, as emphasized in recent studies~\cite{chang2024matters_arxiv2024,he2025vlforgery_arxiv2025}.
To support this goal, AEGIS provides rich multimodal annotations for each video, covering, \textbf{\textit{Semantic-Authenticity Descriptions}}, \textbf{\textit{Motion Features}}, and \textbf{\textit{Low-level Visual Features}}.

(1) \textbf{\textit{Semantic-Authenticity Descriptions}.} 
To capture both high-level semantics and authenticity-related cues, we provide two types of textual descriptions for every video: semantic descriptions and authenticity reasoning descriptions.
For synthetic videos, we directly use the original prompts from the TIP-I2V dataset as semantic descriptions, which specify the intended scene, objects, and actions.
For real videos, where no prompts are available, we extract frame-level embeddings using CLIP~\cite{radford2021clip_icml2021} and apply $k$-means clustering ($k=8$) to identify representative key frames. We then query GPT-4V~\cite{openai2023gpt4v} to generate semantic descriptions summarizing the content of these key frames.
In addition, for both real and synthetic videos, we provide authenticity reasoning descriptions. For each video, we inform GPT-4V of its ground-truth label (real or AI-generated), and prompt it to explain the reasoning behind the label based solely on the visual content. These explanations may highlight temporal smoothness, lighting consistency, or the presence of visual artifacts, offering human-interpretable insights into authenticity cues.

(2) \textbf{\textit{Motion Features.}} 
Realistic motion tends to be temporally smooth and physically coherent, whereas synthetic videos often exhibit subtle artifacts or violations of natural dynamics.
To capture such motion inconsistencies, we extract dense optical flow fields using the RAFT algorithm~\cite{teed2020raft_eccv2020}, enabling fine-grained characterization of frame-to-frame motion patterns.

(3) \textbf{\textit{Low-level Visual Features.}} 
Low-level vision features address subtle yet revealing pixel-level and frequency-domain discrepancies, such as edge sharpness, compression artifacts, overly smooth textures or repetitive patterns and dynamic range variations. 
we compute the 2D Fast Fourier Transform (FFT) of each grayscale key frame and apply Radial Integral Operations (RIO) to summarize frequency energy across orientations.

\subsection{Distinctive Contributions of AEGIS}
The proposed \textbf{AEGIS} dataset advances video authenticity detection by explicitly addressing the challenges posed by hyper-realistic AI-generated videos that closely resemble authentic human-created content.
Unlike existing benchmarks that often include stylized animations or trivially detectable scenarios, AEGIS focuses exclusively on visually nuanced and contextually rich videos, deliberately curated to reflect the complexities of real-world detection tasks.

Furthermore, AEGIS leverages GPT-4o-refined prompts and state-of-the-art proprietary generative models—such as KLing and Sora—to synthesize highly realistic and deceptive forgeries. These, combined with carefully selected authentic samples, form the \textit{Hard Test Set}, a rigorous benchmark designed to evaluate model robustness and generalization under challenging, real-world conditions.

In addition, AEGIS provides ready-to-use multimodal visual cues to support downstream tasks in synthetic video detection and interpretable reasoning, facilitating deeper insight into model behavior and failure cases.
\section{Benchmarking on AEGIS}
In this section, we design evaluation strategies to benchmark the authenticity detection performance of SOTA vision-language models on our AEGIS dataset.
\subsection{Benchmarking Setup}
We evaluate two subsets. (i) An in-domain test set: a subset randomly sampled from the validation set, sharing the same distribution as the training data;
(ii) The Hard Test Set (see Sec. \ref{sec:data_split}): an out-of-domain split explicitly designed to assess model robustness and generalizability on challenging synthetic videos.

\noindent\textbf{Baseline Models.} 
We evaluate two SOTA vision-language models on the AEGIS dataset: Qwen2.5-VL~\cite{bai2025qwen2.5-vl_arxiv2025} and Video-LLaVA~\cite{lin2023videollava_arxiv2023}.
Qwen2.5-VL is a strong general-purpose model with robust multimodal comprehension and competitive performance on video-centric tasks; we consider both its 3B and 7B variants.
Video-LLaVA is a representative auto-regressive transformer model designed to unify image and video understanding within a single framework.

\subsection{Benchmarking Strategies}
To systematically examine model performance under different levels of task conditioning, we implement three evaluation strategies: (i) \textit{\textbf{Zero-shot Inference}}, (ii) \textit{\textbf{Structured Reasoning Prompt}}, and (iii) \textit{\textbf{Low-Rank Adaptation (LoRA) ~\cite{hu2022lora_iclr2022} fine-tuning}}.
To leverage extracted multimodal cues during inference, pre-extracted key frames are fed to the vision-language models using the <image> token format supported by the model interface. 

(1) \textbf{\textit{Zero-shot Inference.}}
We employ a minimal prompt to solicit a binary decision from the model:
\small \textit{“You are an expert in AI-generated-content (AIGC) detection. Given a video, determine whether it is real or AI-generated.”}
This setup evaluates the model’s default capability to perform authenticity detection given only a task description.

(2) \textbf{\textit{Structured Reasoning Prompt.}}  
We construct a multi-step prompt that guides the model through a detailed reasoning process over several visual dimensions, including frequency artifacts, lighting consistency, compression noise, and physical plausibility. Our reasoning-enhanced prompt template is provided in the supplementary link. 

(3) \textbf{\textit{LoRA Fine-tuning.}}  
To explore task-specific adaptation, we fine-tune Qwen2.5-VL ~\cite{bai2025qwen2.5-vl_arxiv2025} on the training set using LoRA ~\cite{hu2022lora_iclr2022} (learning rate $1e^{-4}$, rank 8, 3 epochs). We utilize the widely-used framework llama-factory ~\cite{zheng2024llamafactory_arxiv2024} for effective training. This setting serves to quantify potential gains from lightweight supervision and evaluate model generalization beyond training distribution.

For each setting, we report four metrics:
$Acc_{all}$: overall classification accuracy (from 0 to 1).
$Acc_{real}$: accuracy on authentic videos;.
$Acc_{ai}$: ccuracy on synthetic videos.
Macro-$F1$: Unweighted average F1 score across the two classes.

\subsection{Benchmarking Results}

Experiments conducted on the AEGIS dataset serve two purposes: 
(i) to show that current VLMs struggle with video authenticity evaluation on AEGIS, and (ii) to test whether additional training on AEGIS improves their performance.

\noindent\textbf{AEGIS Reveals Gaps in VLMs Zero-Shot Detection.}
As shown in \autoref{tab:hard-test-set zero-shot}, SOTA models like Qwen2.5-VL~\cite{bai2025qwen2.5-vl_arxiv2025} achieve low synthetic video detection accuracy ( $Acc_{ai}$ from 0.22 to 0.23 ) on the AEGIS Hard Test Set under zero-shot settings. This highlights the substantial gap between existing model capabilities and the high visual fidelity of AEGIS samples.
Furthermore, prompt-based reasoning offers little improvement. 
As illustrated in \autoref{tab:hard-test-set prompt-based}, accuracy further drops from 0.22 to 0.16 when applying direct textual prompts to Qwen2.5-VL 7B. This unexpected performance indicates that conventional prompting strategies fail to capture the nuanced visual and semantic cues characteristic of high-quality forgeries in AEGIS.

\begin{table}[h!]
  \footnotesize    \centering
      \caption{Detection Accuracy on Hard Test Set}
  \vspace{-1.5em}
      \label{tab:hard test set accuracy}
          \subcaption{Zero-shot Inference}
          \vspace{-0.5em}
          \begin{tabular}{ccccccc}
              \hline
\rowcolor[HTML]{EFEFEF}              Model              & $Acc_{all}$ & $Acc_{real}$ & $Acc_{ai}$ & Macro F1 \\ \hline
              Qwen2.5-VL   3B    & 0.52        & 0.80         & 0.23       & 0.48     \\
              Qwen2.5-VL   7B    & 0.59        & 0.89         & 0.22       & 0.52     \\
              Video-LLaVA-HF 7B  & 0.5         & 0.0          & 1.0        & 0.33     \\\hline
          \end{tabular}
          \label{tab:hard-test-set zero-shot}
          \vspace{0.5em}
          \subcaption{Structured Reasoning Prompt}
          \vspace{-0.5em}
          \begin{tabular}{ccccccc}
              \hline
      \rowcolor[HTML]{EFEFEF}        Model              & $Acc_{all}$ & $Acc_{real}$ & $Acc_{ai}$ & Macro F1 \\ \hline
              Qwen2.5-VL   3B    & 0.47        & 0.58         & 0.35       & 0.46     \\
              Qwen2.5-VL   7B    & 0.57        & 0.97         & 0.16       & 0.48     \\
              Video-LLaVA-HF 7B  & 0.46        & 0.29         & 0.63       & 0.45     \\ \hline
          \end{tabular}
          \label{tab:hard-test-set prompt-based}
  \vspace{-1.3em}
  \end{table}

\noindent\textbf{Training on AEGIS boosts authenticity detection.}
Fine-tuning with LoRA yields substantial performance gains on the in-domain test set—for instance, the macro-F1 of Qwen2.5-VL 7B increases from 0.43 to 0.82. However, as shown in \autoref{tab:test-set-after-ft}, improvements on the Hard Test Set remain marginal, with macro-F1 rising only slightly from 0.52 to 0.55. This underscores the persistent challenge of generalization to realistic, high-fidelity forgeries.

The stark contrast between performance on in-domain data and the Hard Test Set underscores the critical generalization challenges posed uniquely by AEGIS. 
Despite targeted fine-tuning, current models still struggle to generalize learned authenticity cues effectively when confronted with subtle and realistic videos deliberately included in the Hard Test Set.

\begin{table}[h!]
\footnotesize \centering
\caption{Detection Accuracy on Two Test Set after Fine-tuning}
\vspace{-0.5em}
\label{tab:test-set-after-ft}
\begin{tabular}{ccccccc}
\hline
\rowcolor[HTML]{EFEFEF} M & T & Eval & $Acc_{\text{all}}$ & $Acc_{\text{real}}$ & $Acc_{\text{ai}}$ & Macro-$F1$ \\
\hline
3B & ID & ZS  & 0.65 & 0.87 & 0.55 & 0.65 \\
7B & ID & ZS  & 0.45 & 0.50 & 0.20 & 0.43 \\
3B & ID & LoRA & 0.78 & 0.35 & 1.00 & {\color[HTML]{009901} 0.69 (+0.04)} \\
7B & ID & LoRA & 0.83 & 0.99 & 0.75 & {\color[HTML]{009901} 0.82 (+0.41)} \\
3B & HT & ZS  & 0.52 & 0.80 & 0.23 & 0.48 \\
7B & HT & ZS  & 0.59 & 0.89 & 0.22 & 0.52 \\
3B & HT & LoRA & 0.56 & 0.14 & 0.97 & {\color[HTML]{FE0000} 0.47 (-0.01)} \\
7B & HT & LoRA & 0.61 & 0.99 & 0.24 & {\color[HTML]{009901} 0.55 (+0.03)} \\
\hline
\end{tabular}

\vspace{0.5em}
  \begin{minipage}{\linewidth}
    \footnotesize
    \textbf{M}: Model size (3B = Qwen2.5-VL-3B, 7B = Qwen2.5-VL-7B); 
    \textbf{T}: Test set (ID = In-domain, HT = Hard test set); 
    \textbf{Eval}: Evaluation Type (ZS = Zero-shot, LoRA = After LoRA fine-tuning).
  \end{minipage}
\end{table}

This limitation strongly indicates the need for future research to explore more advanced and robust fine-tuning or domain adaptation strategies explicitly tailored toward enhancing model generalization to AEGIS-level forgery complexities.
These insights collectively underline the unique value and significant challenge presented by AEGIS, clearly establishing it as a critical resource for advancing robust, realistic, and highly generalizable AI-generated video detection research.

\section{Conclusion}
In this work, we presented {AEGIS}, a novel large-scale video authenticity benchmark explicitly targeting sophisticated AI-generated videos. 
Unlike existing datasets, AEGIS prioritizes hyper-realistic scenarios and excludes simplistic or easily detectable samples, significantly enhancing detection complexity and realism. 
Through rigorous data filtering, strategic dataset partitioning, and the inclusion of deceptive samples from advancing generative models (e.g., Sora, KLing), AEGIS raises the bar for synthetic video detection. 
Experimental evaluations reveal that even SOTA vision-language models struggle to generalize in zero-shot settings, particularly on the Hard Test Set.
Furthermore, AEGIS offers multi-dimensional visual cues with rich multimodal annotations. These not only support downstream detection tasks but also facilitate interpretable reasoning, enabling finer-grained analysis of model failures and decision boundaries.
We believe AEGIS establishes a foundational shift in synthetic video detection research by providing a challenging, diverse, and explainability-oriented benchmark, essential for the development of robust and trustworthy multimodal AI systems.

\section*{Acknowledgements}
This research is supported by the National Research Foundation, Singapore 
under its AI Singapore Programme (AISG Award No: AISG3-RP-2024-033).

\newpage
\bibliographystyle{ACM-Reference-Format}
\balance
\bibliography{sample-base}

\end{document}